# Simultaneously Calibration of Multi Hand-Eye Robot System Based on Graph

Zishun Zhou, Liping Ma, Xilong Liu, Zhiqiang Cao, *Senior Member, IEEE,*
Junzhi Yu, *Fellow, IEEE*

*Abstract*—Precise calibration is the basis for the vision-guided robot system to achieve high-precision operations. Systems with multiple eyes (cameras) and multiple hands (robots) are particularly sensitive to calibration errors, such as micro-assembly systems. Most existing methods focus on the calibration of a single unit of the whole system, such as poses between hand and eye, or between two hands. These methods can be used to determine the relative pose between each unit, but the serialized incremental calibration strategy cannot avoid the problem of error accumulation in a large-scale system. Instead of focusing on a single unit, this paper models the multi-eye and multi-hand system calibration problem as a graph and proposes a method based on the minimum spanning tree and graph optimization. This method can automatically plan the serialized optimal calibration strategy in accordance with the system settings to get coarse calibration results initially. Then, with these initial values, the closed-loop constraints are introduced to carry out global optimization. Simulation experiments demonstrate the performance of the proposed algorithm under different noises and various hand-eye configurations. In addition, experiments on real robot systems are presented to further verify the proposed method.

## I. INTRODUCTION

The hand-eye robot systems are used in a wild range. Compared with the single-eye and one-hand systems, the multi-eye and multi-hand systems (MEMHS) can complete flexible and various tasks through cooperation and association, which is a research hotspot at present. Many difficult tasks for a single robot can be easily accomplished by multi-hand-eye robot systems [1], and the MEMHS has been widely used in various fields. For example, Tianjin University has designed a multi-robot mirror milling system to machining large skin parts. In order to ensure that the milling accuracy is less than 0.03mm, two robotic arms need to move synchronously to mill and support both sides of the metal at the same time. This system has high requirements for calibration accuracy [2]. Institute of Automation, Chinese Academy of Sciences has designed a micro-assembly system for miniature photoelectric lens manufacturing based on 3 cameras and 6 manipulators. High-precision calibration is the cornerstone of the 21-degree-of-freedom (DoF) control and 4.9μm assembly accuracy of the system [3]. Other applications also include dexterous manipulation [4], grasping [5], and robot-guided 3-D feature recognition [6], which play an important role in intelligent manufacturing. But due to the difficulty of high precision calibration, the multi-eyes and multi-hands system cannot achieve high precision collaboration at low cost. For example, Boeing is straggling with its new multi-robot assembly system, which leads to the delayed delivery of its new Airplane [7].

Manuscript received October 2022; revised Month xx, xxxx; accepted Month x, xxxx. This work was supported by the National Natural Science Foundation of China under Grants 61973302 2020-01-01~2023-12-31, National Defense Basic Research Program Key Projects under Grant JCKY2020208B027, the CIE-Tencent Robotics X Rhino-Bird Focused Research Program No. 2022-07 and Beijing Natural Science Foundation (2022MQ05). (Corresponding author: Liping Ma)

Zishun Zhou is a student intern under the supervision of Xilong Liu, the Institute of Automation, Chinese Academy of Sciences, Beijing 100190, China, and he is also a postgraduate student at the College of Electrical Engineering and Computer Science, Australian National University, Canberra 2601, Australia

Liping Ma, Xilong Liu, and Zhiqiang Cao are with the State Key Laboratory of Management and Control for Complex Systems, Institute of Automation, Chinese Academy of Sciences, Beijing 100190, China, and also with the School of Artificial Intelligence, University of Chinese Academy of Sciences, Beijing 100049, China.

Junzhi Yu is with the State Key Laboratory for Turbulence and Complex System, Department of Advanced Manufacturing and Robotics, BIC-ESAT, College of Engineering, Peking University, Beijing 100871, China.

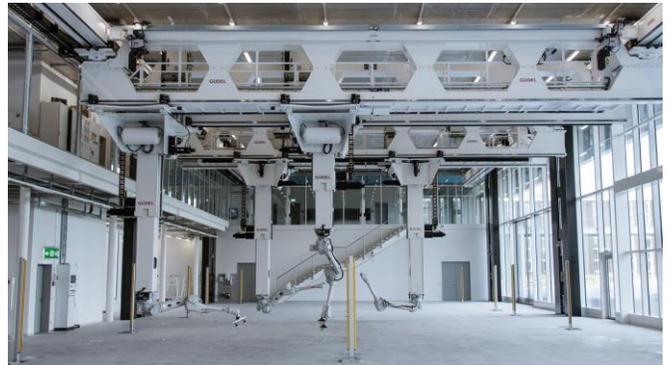

Fig. 1 The ETH Zurich's Robotic Fabrication Laboratory (RFL) is a MEMHS. It consists of four industrial robots with six axes each, suspended from an overhead gantry system, which allows the robot end-effectors to be positioned at any desired angle within a spacious work area measuring 45 by 17 by 6 meters.

In this paper, the MEMHS refers to a system that consists of at least three robots and three cameras. Cameras fixed in the workspace are called eye-to-hand cameras and those mounted on the robotic arms are called eye-in-hand cameras. The parameters to be calibrated in the MEMHS include the kinematic parameters of the robotic arms, the intrinsic parameters of the cameras, and the relative poses between units (e.g., cameras-robots, robots-robots, cameras-cameras). The calibration of the kinematic parameters of the robot itself and the intrinsic parameters of the cameras have been widely studied [8-12], and are effective enough for most applications. The calibration of the extrinsic pose matrices between the system's units is comparably difficult, especially in tasks with high precision requirements, which is a major bottleneck restricting the wide application of MEMHS.

### A. Related Works

Robot hand-eye calibration is a classic and basic problem for single robot extrinsic calibration. It concentrates on solving the pose matrices between the robot base or end flange and the sensor frame, which is the famous $AX = XB$ and $AX = YB$ problem, where $X$ and $Y$ are unknown constant transformation matrices [13]. Tsai et al. [14] use a series of matrix geometries and properties of homogeneous matrices to compute the rotation and translation of the hand-eye relationship. However, since the parameters are estimated in a two-stage process, the estimation error from the first stage will propagate to the second stage. Park et al. [15] proposed a method to calculate $AX = XB$ in a one-stage process. This method uses the Lie group theory to solve the rotation and translation of the hand-eye equation simultaneously. Ernst et al. [16] proposed a method calibrating tool-flange and robot-world simultaneously with a least-squares approach, that achieved more accurate results and required fewer calibration stations.

In previous studies on multi-robot calibration problems, researchers mainly focused on the dual robot system, which is the well-known $AXB = YCZ$ problem. Where $X$, $Y$, and $Z$ are all unknown constant matrices, representing the transformation between a base frame and another base frame, camera frame and end flange, another end flange and tool frame respectively. By solving the dual robot calibration problem, the multi-robot calibration problem could be easily solved by decomposing the multi-robot system into a series of independent dual robots groups, or hand-eye robot units, then applying the dual robot or hand-eye calibration method on each group [17]. Namely, step-by-step calibrate an open-loop chain that is composed of all units. Wang et al. [1] proposed a method to simultaneously calibrate the parameters of a dual robot system using Lie algebra and the non-linear optimization method. The initial value is obtained by solving two $AX = XB$ problems. Wu et al. [18] also used the nonlinear optimization method of Lie algebra to calibrate the dual-hand-eye robot system. To obtain a good initial value, they use quaternion to calculate all the unknown pose matrices synchronously in avoiding error propagation before performing global adjustments. Fu et al. [19] use a dual quaternion-based analytical solution for dual robot calibration, and solvability analysis is provided to prove the robustness against noise. Wang [17] used the Kronecker product and iteration to solve the problem, proofing that their cost function is strictly convex. Ma et al. [20] proposed a hybrid approach that combines $AXB = YCZ$ solvers from Wang [1] with probabilistic methodology, which requires fewer data without losing accuracy. It considered the noise level of different sensor, and information matrices has been added during the optimization process, however, it only focuses on the dual robot scenario, which only consists of one loop, and it may deteriorate quickly as the noise level increase.

Less attention has been paid to the calibration sequence of different units and the accumulative error problem in actual operations, even though it has a vital influence on multi-robot co-manipulation. Many scholars have studied the calibration sequence of MEMHS, but most of these methods are just suitable for specific scenarios and lack flexibility and scalability. Ruan et al. [21] produced a 3D camera multi-robot system and designed a non-contact calibration strategy for this system. This approach calibrates their cooperative grinding workstation accurately, but it uses a special geometric relationship of their system, which may be difficult to transplant to another system. Shen et al. [3] designed a micro-assembly platform with 3 cameras and 6 manipulators to achieve 21-DoF assembly tasks. This platform is calibrated step by step using the single unit calibration method proposed in [22]. This approach also only supports their specially designed system. Yu et al. [23] proposed a general method to calibrate both extrinsic matrices between robots and robot kinematic parameters. This method uses the product of exponentials (POE) formula to module robot motion and determine the unknown parameters. This method is a general method for multi-robot calibration, while kinematic parameters can be determined simultaneously. However, this method did not consider sensor precision difference, and the selection of world coordinates may influence the calibration accuracy. Due to the complexity of offline calibration, some pose compensation-based calibration methods are proposed. Stadelmann et al. [24] proposed an end-effector pose correction method for large-scale multi-robot systems. They use Indoor Global Positioning System (iGPS) and IMU to track the end-effector pose, thereby achieving high accuracy calibration. With their sensor fusion algorithm, the end-effector position error reduces to 0.1mm on average. Maghami et al. [25] use deep neural networks for multi-robot calibration. Measure the pose of multiple robots at different locations and record the joint angle input simultaneously. Then trained a fully connected network for error compensation.

### B. Contributions

In this article, we mainly focus on the general calibration process of MEMHS, and special attention has been provided to finding the optimal calibration path with the highest sensor accuracy, which is the equilibrium point of efficiency and accuracy. This paper models the MEMHS calibration problem as a graph and proposes a minimum spanning tree and graph optimization-based calibration strategy. Firstly, optimal open-loop chains are selected for local calibration to get a set of initialization parameters. Then, closed loops covering all units are selected for global adjusting. The main contributions of this paper are as follows:

1) A general multi-eye multi-hand system (MEMHS) calibration model based on minimum spanning tree and graph optimization is presented, and the optimal calibration principle is proposed.

2) According to the presented system model, a numerical solution of nonlinear calibration equations is given, and the corresponding algorithm is developed.
3) A standard calibration procedure for MEMHS is proposed, which is comprehensively tested on both simulated and real MEMHS.

The paper is organized as follows: In section II, the calibration problem is modeled. In section III, details of the proposed calibration method are introduced. Simulation and real robot verification of the calibration method are presented in Section IV. Finally, Section V concludes this article.

## II. PROBLEM STATEMENT

This paper is devoted to the calibration of a system that consists of multiple cameras and robotic arms. On the premise that the cameras' intrinsic parameters and the kinematic parameters of the robots themselves are well calibrated, given a set of robot joint states and camera measurement results, the pose between any two units can be calibrated.

Denote $N$ robotic arms, $M$ eye-in-hand cameras, and $L$ eye-to-hand Cameras as a MEMHS. The relative pose between the bases of the $N$ robots remain constant, $M$ eye-in-hand cameras are mounted on the bodies of robotic arms and each of their pose is constant to a certain joint. The poses of $L$ eye-to-hand Cameras are fixed to the base of $N$ robots. The base coordinate system of the $i$-th robotic arm is denoted as $\{R_i\}$. The tool coordinate system is represented by $\{H_i\}$, the coordinate system of the $j$-th eye-in-hand camera is represented by $\{E_j\}$, and the coordinate system of the $k$-th eye-to-hand camera is represented by $\{C_k\}$, where $i = 1,2 \ldots \ldots N, j = 1,2 \ldots \ldots M, k = 1,2, \ldots \ldots L$. Poses between devices are changing with robots' joint angles. Mathematically, this can be described as $\forall A, B \in \{R_i\}, \{E_j\}, \{H_i\}, \{C_k\}$. the transformation matrix from $\{A\}$ to $\{B\}$ is denoted as ${}^B T_A$, where ${}^B T_A \in SE(3)$.

Fig. 2(a) shows a MEMHS, which consists of three robots, three eye-in-hand cameras, and one eye-to-hand camera. The solid line in the figure represents the measurable pose matrices, the blue one indicates the matrices which can be obtained from the joint angles with corresponding kinematic parameters of the robot; The red one indicates the matrices which can be obtained from cameras with overlapping fields of view. The dashed line represents the constant pose matrices that need to be calibrated.

## III. METHOD

Taking each coordinate system as a vertex and the pose relationship between two coordinate systems as an edge, the MEMHS can be modeled as a complete graph, which is represented as follows.

$$G(V, E, \Phi) \quad (1)$$

where $V$ is the vertex set of $G$, $E$ is the edge set of $G$, and $\Phi$ is the weight set of the corresponding edge. Calling the MEMHS complete graph MCG for short. Each edge corresponds to a translation matrix, where some edges are constant matrices, and some are functions of joint angles, and its accuracy is related to weight $\Phi$. The calibration problem can then be formulated as determining a set of parameters such that for a given system joint angles state, the matrices corresponding to all edges can be calculated. If MCG has $n$ vertices in total, then there are $n(n-1)$ edges in total, but only $n-1$ of them are independent, since MCG has an important property that along each of a close ring, the product of the corresponding matrices should be an identity matrix. Therefore, all spanning trees of the graph are mathematically equivalent, namely, any given spanning tree can restore the system graph. Different from SLAM (Simultaneous localization and mapping) problem, the calibration problem requires human labor forces. The $n(o^2)$ time complexity may not be high for computers when n is not very large, but it could be extremely large for human force. Therefore, calibrating the $n-1$ best edges will balance the accuracy and the efficiency. The minimum spanning tree is the best tree for preliminary calibration. Then, according to the spanning tree equivalence principle and preliminary calibration results, a set of closed loops of MCG are selected for global optimization based on the close-loop-identity property.

Take the MEMHS shown in Fig. 2(a) as an example, assuming the kinematic parameters of the robots and intrinsic parameters of the cameras are known, the MCG is shown in Fig. 2(b). Red and black edges represent the translation matrices and can be directly measured according to the kinematic model of the robot or cameras; the blue edge represents the edge corresponding to the constant matrices.

### A. edge evaluation

The measurement accuracy mainly depends on the accuracy of the sensor itself, and the sensitivity of the measurement process. The former includes the pixel density of the camera, the encoder accuracy of the robotic arm, etc. The latter includes the distance between the calibration object and the camera, etc. In the MCG, the former is related to vertices while the latter is related to edges. Therefore, it is necessary to set different weights for vertices and edges with different precisions to achieve better measurement results. Let the weight of vertex $i$ be $\eta_i \in \mathbf{R}^+$, the vertices with higher measurement accuracy will have larger weights. From the perspective of practical engineering, $\eta_i$ is determined by an empirical formula including the influence mentioned above. For the case where the covariance matrices are known, SVD decomposition is performed on the corresponding covariance matrix, and its largest eigenvalue is taken as $\eta_i$. Then convert the weight of the vertex into the weight of the edge. Let the currently measured vertices be $i, j$, and let the edge weight be $\Phi_{ij}$. Additionally, for the sensitivity of the measurement process, it is necessary to add preset weights on the edge directly, but this error is difficult to obtain before measurement, so it is simply represented by $d_{ij}$ here. $d_{ij}$ should be constant zero in most cases, only when the physical distance of two vertices is too large, or existing obstacles or difficulties affect the measurement in the given edge, the $d_{ij}$ need to be added. The final edge weights can be expressed as follows.

$$\Phi_{ij} = \frac{1}{\log\left(n_{ij}\left(\frac{1}{\eta_i} + \frac{1}{\eta_j} + d_{ij}\right) + 1\right) + 1} \quad (2)$$

where $n_{ij}$ represents the measurement times of edge between vertices $i, j$. We assume that in unbiased measurements the accuracy of the data measurement is more important than the

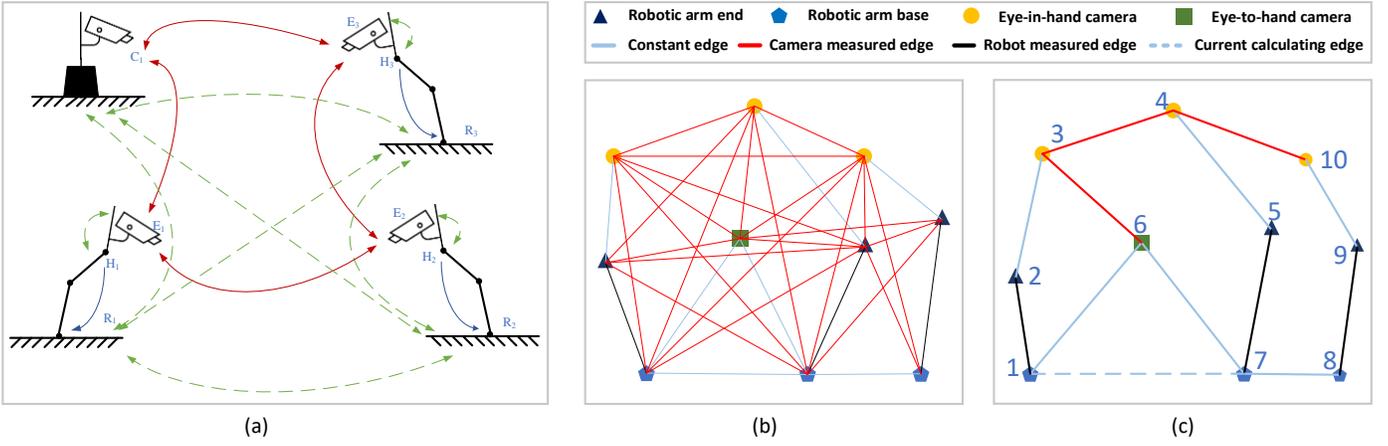

Fig. 2 (a) Example of MEMHS, which consists of three robots, three eye-in-hand cameras, and one eye-to-hand camera. (b) The corresponding MCG (MEMHS complete graph) of the MEMHS shown in (a). (c) The optimal calibration loop diagram of the MEMHS shown in (a).

quantity, so the weight function uses a logarithmic term to limit the number of measurements. The number of measurements $n_{ij}$ for edges that cannot be directly measured or computed by hand-eye calibration (such as transportation matrices between two robot bases) is 0. Weights are assigned to different edges, and the minimum weight constraint and spanning tree criterion are introduced to simplify the graph to find the optimal calibration path. This paper uses Prim's algorithm [26] to generate the minimum spanning tree, since the time complexity of Prim's algorithm is $o(n^2)$, where $n$ is the number of vertices, and its time complexity is only related to the number of vertices.

### B. Calibration of edge

The calibration problem of the MEMHS can be divided into a series of calibration tasks between two vertices. Calibration methods between a pair of units have been well studied. For the calibration between fixed cameras, if the two cameras have enough overlapping fields of view, the problem becomes a dual camera calibration problem. The intrinsic parameters of each camera are obtained separately [8] [9]. Then by placing the calibration board in the public field of view, and using the PnP algorithm [27] [28] to measure the pose between the cameras and the calibration board, the pose relation of the two camera units can be calculated.

The calibration problem between the camera and the end of the robot, and between the camera and the base of the robot can be modeled in the form of eye-in-hand and eye-to-hand problems. The calibration equation has the form of $AX = XB$ [14] [15]. where $X$ represents the matrix to be solved. The calibration can be completed by placing a calibration board in a fixed position in the working area or at the end of the robot arm.

### C. Global optimization and Iterative calibration

The main idea of the iterative calibration stage is to construct a closed-loop optimization error function to eliminate the error to the greatest extent. The calibration is done by adjusting the pose matrix $T_i$ between multiple units. If an edge is added between any two non-directly connected vertices in the minimum spanning tree, a ring can be formed, and the matrix obtained by multiplying the corresponding matrices on the ring in order should result in the identity matrix $I$. However, due to the existence of the actual error, the matrix is not an identity matrix after continuous multiplication, causing the coordinates of the same point to be different. Therefore, the error function can be established. The calibration error function has the following form.

$$e = \frac{1}{2} || \sum_{k=0}^{S_{All}} \omega_k \sum_{j=0}^{M} \left( \prod_{i=1}^{N} ({}^{i-1}A_{ijk}^{i-1} X_{ik}) \right)^N P_{jk} - {}^0 P_{jk} ||^2 \quad (3)$$

where $A_i, X_i \in SE(3)$, $\omega_k$ is the weight for optimization, and $P$ is a set of measured points. The accumulated item $i$ represents the number of edges passed on a loop, $A_i$ represents the measurable matrix, $X_i$ represents the matrix that requires optimization, and there will be $N$ measurable and optimizable matrix in the loop $k$, since the loop may be measured multiple times, all the measurements result should be summed. $M$ represents the measurement times of the current loop. weights need to be set when calculating the error, which is represented by $\omega_k$. Where $\omega_k$ is equal to the sum of the inverses of all edge weights in the loop.

$$\omega_k = \sum_{i=0}^{N} \frac{1}{\Phi_i} \quad (4)$$

Finally, the total error of all loops is comprehensively considered, which is represented by the accumulated term $k$.

According to the error model, the optimization aim can be expressed as follows.

$$X = [X_{11}, X_{12}, \ldots, X_{SN}] = \arg \min_X \frac{1}{2} || f(X) ||^2 \quad (5)$$

### D. Determining optimal calibration loop

When handling the calibration task of the MEMHS into a weighted graph, calculating all the errors of all loops is redundant, and its time complexity is too high. Secondly, it is difficult to obtain direct calibration values between many vertices. Therefore, the optimal calibration loop needs to be found to reduce the complexity and improve the calibration accuracy. The optimal calibration loop has the following characteristics. It contains at least one edge requiring calibration; contains at least one measurable edge; the loop has the shortest weight distance, and the weight of each edge on the loop uses the weight determined above. An optimal calibration loop algorithm based on Dijkstra's algorithm is proposed. Fig. 2(c) illustrates the current calculation of the pose matrix for the edge connecting vertices 1 and 7, which is depicted as a dotted line. To determine the optimal calibration loop, path planning is performed on the MCG. The main purpose of processing is to delete some vertices, thereby reducing the amount of calculation and searching for the correct path. To prevent a loop from being unable to form, firstly, delete the edge directly connecting vertices 1 and 7. Secondly, delete edges with too large distances or no connections. As shown in Fig. 2(c). The relationship between vertex 2 and vertex 6 (corresponding to the coordinate system of the end of the robot and the fixed camera coordinate system) is difficult to measure directly without introducing additional instruments such as laser trackers, therefore, it needs to be deleted before path planning.

The next step is searching for the shortest path. Dijkstra's algorithm [29] is used in this step. Then after deleting redundant calibration loop, the error function has the following form.

$$e = \frac{1}{2}||f(X_{11}, X_{12}, \ldots, X_{SN})||^2 \quad (6)$$

$$f(X_{11}, X_{12}, \ldots, X_{SN}) = \sum_{k=0}^{S} \omega_k \sum_{j=0}^{M} \left( \prod_{i=1}^{N} \left( {}^{i-1}A_{ijk} {}^{i-1}X_{ik} \right) {}^{N}P_{jk} - {}^{0}P_{jk} \right) \quad (7)$$

### E. Error Correction Model and Correction Method

An error model based on the determined optimal calibration loop is obtained. Numerical optimization begins with an initial value and then involves the computation of derivatives by introducing disturbances to the error model and repeating the process iteratively, by adding disturbance and computing the derivative of disturbance, the complexity of computing Jacobian matrices is reduced. The direction of descent is determined by computing the linear extrema on the tangent plane. Recent research on the 3D bundle adjustment problem shows the highly non-convex nature of graph optimization [30], Nevertheless, thanks to the basically accurate initial values generated by the first stage, the chance of falling into local minima is significantly reduced.

Since the translation matrix $X_{ki}$ belongs to the Lie group $SE(3)$, it can be exponentially mapped by its Lie algebra, where $\wedge$ is the anti-symmetric symbol, transforming the 3x1 vector into a 3x3 anti-symmetric matrix [31].

$$\xi^\wedge = \begin{bmatrix} \boldsymbol{\phi}^\wedge & \boldsymbol{\rho} \\ \mathbf{0} & 0 \end{bmatrix} \quad (8)$$

Then the calibration error function $f(X_{11}, X_{12}, \ldots, X_{SN})$ equals to $g(\xi_{11}, \xi_{12}, \ldots, \xi_{SN})$.

$$f(X_{11}, X_{12}, \ldots, X_{SN}) = g(\xi_{11}, \xi_{12}, \ldots, \xi_{SN}) = \sum_{k=0}^{S} \omega_k \sum_{j=0}^{M} \left( \prod_{i=1}^{N} \left( {}^{i-1}A_{ijk} e^{\xi_{ki}^\wedge} \right) {}^{N}P_{jk} - {}^{0}P_{jk} \right) \quad (9)$$

After adding disturbance, the current goal is to find the disturbance vector $\delta\xi$ such that $\frac{1}{2}||g(\xi + \delta\xi)||^2$ reaches the minimum.

$$\delta\xi = \arg\min_{\delta\xi} \frac{1}{2}||g(\xi) + J(\xi)\delta\xi||^2 \quad (10)$$

where

$$\xi = [\xi_{11}, \xi_{12}, \ldots, \xi_{SN}] \quad (11)$$

$$\delta\xi = [\delta\xi_{11}, \delta\xi_{12}, \ldots, \delta\xi_{SN}] \quad (12)$$

$J(\xi)$ is the Jacobian matrix of $g(\xi)$. Then, add disturbance to the error function.

$$g(\xi + \delta\xi) = \sum_{k=0}^{S} \omega_k \sum_{j=0}^{M} \left( \prod_{i=1}^{N} \left( {}^{i-1}A_{ijk} e^{\delta\xi_{ki}^\wedge} e^{\xi_{ki}^\wedge} \right) {}^{N}P_{jk} - {}^{0}P_{jk} \right) \quad (13)$$

Then performing a first-order Taylor expansion on the error function.

$$g(\xi + \delta\xi) \approx g(\xi) + J(\xi) \cdot \delta\xi \quad (14)$$

Then equation (10) can be expanded as follows.

$$\delta\xi = \arg\min_{\delta\xi} \frac{1}{2} \left( ||g(\xi)||^2 + 2g(\xi)J(\xi)\delta\xi + (\delta\xi)^T J(\xi)^T J(\xi)(\delta\xi) \right) \quad (15)$$

Find the derivative of the above equation for $\delta\xi$ and set it to zero.

$$2J(\xi)g(\xi) + 2J(\xi)^T J(\xi)\delta\xi = 0 \quad (16)$$

Then,

$$\delta\xi = \left( J(\xi)^T J(\xi) \right)^{-1} J(\xi)^T g(\xi) \quad (17)$$

The remaining problem is to find the Jacobian matrix $J(\xi)$, which is the key in this process. $J(\xi)$ can be represented as follows.

$$J(\xi) = \frac{\partial g}{\partial \xi} = \left[ \frac{\partial g}{\partial \delta\xi_{11}}, \frac{\partial g}{\partial \delta\xi_{12}}, \ldots, \frac{\partial g}{\partial \delta\xi_{SN}} \right] \quad (18)$$

According to the definition of derivatives, take the partial derivative of $\frac{\partial g}{\partial \xi_{KI}}$ as an example.

$$\frac{\partial g}{\partial \xi_{KI}} = \frac{g(\xi_{11}, \ldots, \xi_{KI} + \delta\xi_{KI}, \ldots, \xi_{SN}) - g(\xi_{11}, \ldots, \xi_{KI}, \ldots, \xi_{SN})}{(\xi_{KI} + \delta\xi_{KI}) - (\xi_{KI})} \quad (19)$$

$$= \lim_{\delta\xi_{KI} \to 0} \frac{\omega_K \sum_{j=0}^{M} \left( \prod_{i=1}^{I-1} \left( {}^{i-1}A_{ijK} e^{\xi_{Ki}^\wedge} \right) \left( e^{\delta\xi_{Ki}^\wedge} - I \right) e^{\xi_{Ki}^\wedge} \prod_{i=I+1}^{N} \left( {}^{i-1}A_{ijK} e^{\xi_{Ki}^\wedge} \right) {}^{N}P_{jK} - {}^{0}P_{jK} \right)}{\delta\xi_{KI}}$$

By applying Taylor expansions at $\delta\xi \to 0$, $\lim_{\delta\xi \to 0} e^{\delta\xi^\wedge} \approx I + \delta\xi^\wedge$. Then equation (19) is expanded follows.

$$\frac{\partial g}{\partial \xi_{KI}} = \lim_{\delta\xi_{KI} \to 0} \frac{\omega_K \sum_{j=0}^{M} \left( \prod_{i=1}^{I-1} \left( {}^{i-1}A_{ijK} e^{\xi_{Ki}^\wedge} \right) (\delta\xi_{KI}^\wedge) e^{\xi_{Ki}^\wedge} \prod_{i=I+1}^{N} \left( {}^{i-1}A_{ijK} e^{\xi_{Ki}^\wedge} \right) {}^{N}P_{jK} - {}^{0}P_{jK} \right)}{\delta\xi_{KI}} \quad (20)$$

Let $\prod_{i=1}^{I-1} \left( {}^{i-1}A_{ijK} e^{\xi_{Ki}^\wedge} \right) = {}^{0}T_{IjK}$, and $\prod_{i=I+1}^{N} \left( {}^{i-1}A_{ijK} e^{\xi_{Ki}^\wedge} \right) = {}^{I}T_{NjK}$, where the rotation component is ${}^{0}R_{IjK}$, ${}^{I}R_{NjK}$, and the translation component is ${}^{0}t_{IjK}$, ${}^{t}t_{NjK}$ respectively.

Then $\frac{\partial g}{\partial \xi_{KI}}$ can be represented as follows.

$$= \lim_{\delta\xi \to 0} \frac{w_K \sum_{j=0}^{M} \left( {}^{0}T_{IjK} \delta\xi_{KI}^\wedge {}^{I}T_{NjK} {}^{N}P_{jK} - {}^{0}P_{jK} \right)}{\delta\xi_{KI}}$$

$$= \lim_{\delta\xi_{KI} \to 0} \frac{w_K \sum_{j=0}^{M} \left( {}^{0}T_{IjK} \begin{bmatrix} \delta\phi_{KI}^\wedge & \delta\rho_{KI} \\ 0 & 0 \end{bmatrix} {}^{I}T_{NjK} {}^{N}P_{jK} - {}^{0}P_{jK} \right)}{\delta\xi_{KI}}$$

$$= \lim_{\delta\xi_{KI} \to 0} w_K \sum_{j=0}^{M} \begin{bmatrix} {}^{0}R_{IjK} & -{}^{0}R_{IjK} \left( {}^{I}R_{NjK} {}^{N}P_{jK} + {}^{I}t_{NjK} \right)^\wedge \\ 0 & 0 \end{bmatrix}$$

(21) Which is a 4 rows 6 columns matrix. The disturbance vector of the $l$-th iteration is obtained, which is defined as $\delta\xi^l$. Then the new $\xi$ can be obtained.

$$\xi^{l+1} = \xi^l + \delta\xi^l \quad (22)$$

After getting $\xi^{l+1}$, we can get $\delta\xi^{l+1}$ through the above method and complete the next iteration. Through continuous iteration, all vertices to be optimized can be finally solved. The algorithm can be summarized as follows.

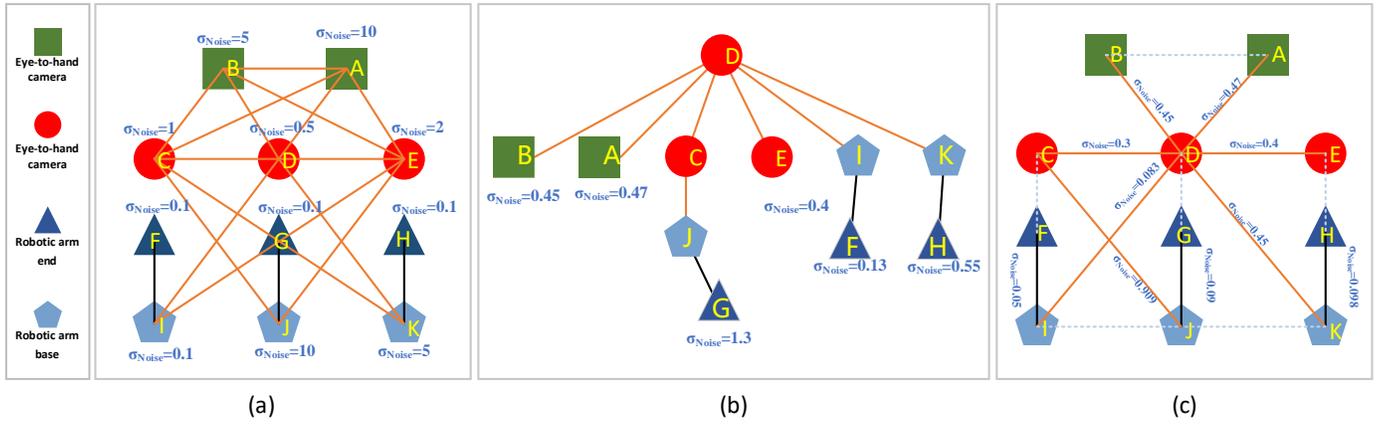

Fig. 3 A simulated MEMHS consisting of three robotic arms and five cameras, including 3 eye-in-hand cameras and 2 eye-to-hand cameras. (a) The MCG of the MEMHS. (b) The expanded minimum spanning tree of the simulated MEMHS. (c) The optimal calibration loop of the simulated MEMHS.

**Algorithm 1.** Calculate the coordinate transformation matrix

**Input:** point $p$ in vertices $i$ and, its representation in vertices $j$, which is $^jP_i$

**Output:** Pose matrix between vertex $i$ and vertex $j$

1. $\delta\xi^A = [\delta\xi_1^\wedge, \delta\xi_2^\wedge, ..., \delta\xi_N^\wedge]$
2. **While** $|\delta\xi^A| > \varepsilon$
3.     calculate $\frac{g(\xi_1^\wedge + \delta\xi_1^\wedge, ..., \xi_N^\wedge + \delta\xi_N^\wedge)}{\delta\xi_i^\wedge}$ based on (21)
4.     calculate $\delta\xi^A$ based on (17)
5.     $\xi_i^\wedge = \xi_i^\wedge + \delta\xi_i^\wedge$
6. **End while**
7. $^iX_j = e^{\xi_i^\wedge}$

## IV. EXPERIMENTS AND RESULT

### A. Simulation

#### 1) Optimal Calibration Loop Simulation

In this section, a simulated system consisting of three robotic arms and five cameras is calibrated. In this system, three cameras are placed at the end of the robot (eye-in-hand), and the remaining two cameras are fixedly installed in the workspace (eye-to-hand). Measurement values with different noise distributions are added. The optimal calibration loop and the results calibrated by the optimal calibration loop are recorded for testing. The MCG of this system and the measurement noise of different edges are shown in Fig. 3(a).

We manually set the weights for different vertices as shown on each vertex of the MCG. Then, convert the weight of each vertex to the weight of corresponding edges, and apply minimum spanning tree strategy to generate the optimal calibration sequences. The optimal calibration tree is shown in Fig. 3(b), where Point D is the root vertex of the generated optimal calibration tree, and the sum of weight for each path is also shown in the figure. The calibration sequence uses the depth-first algorithm to get the initial calibration result of the MEMHS. Finally, the optimal calibration loop obtained with Dijkstra's algorithm method is shown in Fig. 3(c) with each edge weight shown and TABLE I. Closed-loop optimization is conducted on these loops. The ground truth and calibration results between each vertex are shown in TABLE II.

TABLE I
OPTIMAL CALIBRATION LOOP

| Loops |
|---|
| $B \to A \to D \to B$ |
| $C \to D \to I \to F \to C$ |
| $D \to K \to H \to E \to D$ |
| $C \to D \to G \to J \to C$ |

#### 2) Large-scale MEMHS simulation

In this experiment, 25 robotic arms were randomly placed in the workspace, and 25 cameras were randomly mounted at the end of the robotic arm or placed in fixed positions within the workspace. The fully connected weighted graph described above was obtained. Record the robot base, end in 100 poses, as well as camera poses as true values. Then add multiple Gaussian noises with zero means to these poses as simulation measurements.

In this simulation experiment, the proposed method is used to calibrate the above system, and for comparison, some random spanning trees and random calibration loops are generated from the MCG to verify the effectiveness of the proposed algorithm. The MEMHS was initially calibrated with both our method calculated spanning tree and random spanning tree under different noise levels. Then, the average error in each iterative step of the closed-loop optimization stage with loops generated with our method is measured and recorded shown in Fig. 4. For comparison, we also perform closed-loop optimization on randomly generated loops and record the average error at each iterative step for all random calibration loops, the error at each iterative step for the loop with the highest error, and the average error at each iterative step for all loops from the MCG. The simulation results are shown in TABLE III and Fig. 4.

The robustness of the proposed method under large measurement errors is demonstrated in Fig. 5. Our proposed method can automatically select the path with the lowest noise and edges with large errors are not included in the closed-loop optimization, thereby minimizing the impact of large measurement error edges.

TABLE II
THE GROUND TRUTH AND RESULT OF PROPOSED CALIBRATION METHOD (UNIT: MM)

| | Rotation Ground Truth | Rotation Calibration | Rotation Error (rad) | Translation Ground Truth | Translation Calibration | Translation Error (mm) |
|---|---|---|---|---|---|---|
| $^B T_A$ | (0,0,0) | (0.0932,0.0423,0.0312) | (0.0932,0.0423,0.0312) | (1000,0,0) | (996.177,-1.744,-2.7905) | (3.822,-1.744,-2.7905) |
| $^C T_F$ | (1,0,0) | (0.9981,-0.0067,-0.0738) | (0.0019,-0.0067,-0.0738) | (100,0,0) | (102.49,-2.9798,1.7161) | (2.49,-2.9798,1.7161) |
| $^D T_G$ | (0,1,0) | (0.006,1.0055,0.0025) | (0.006,0.0055,0.0025) | (0,100,0) | (-0.5911,100.1840,-0.1829) | (-0.5911,0.1840,-0.1829) |
| $^E T_H$ | (1,1,0) | (0.9975,1.0022,0.0002) | (0.0024,0.0022,0.0002) | (0,0,100) | (0.2827,0.2966,100.9590) | (0.2827,0.2966,0.9590) |

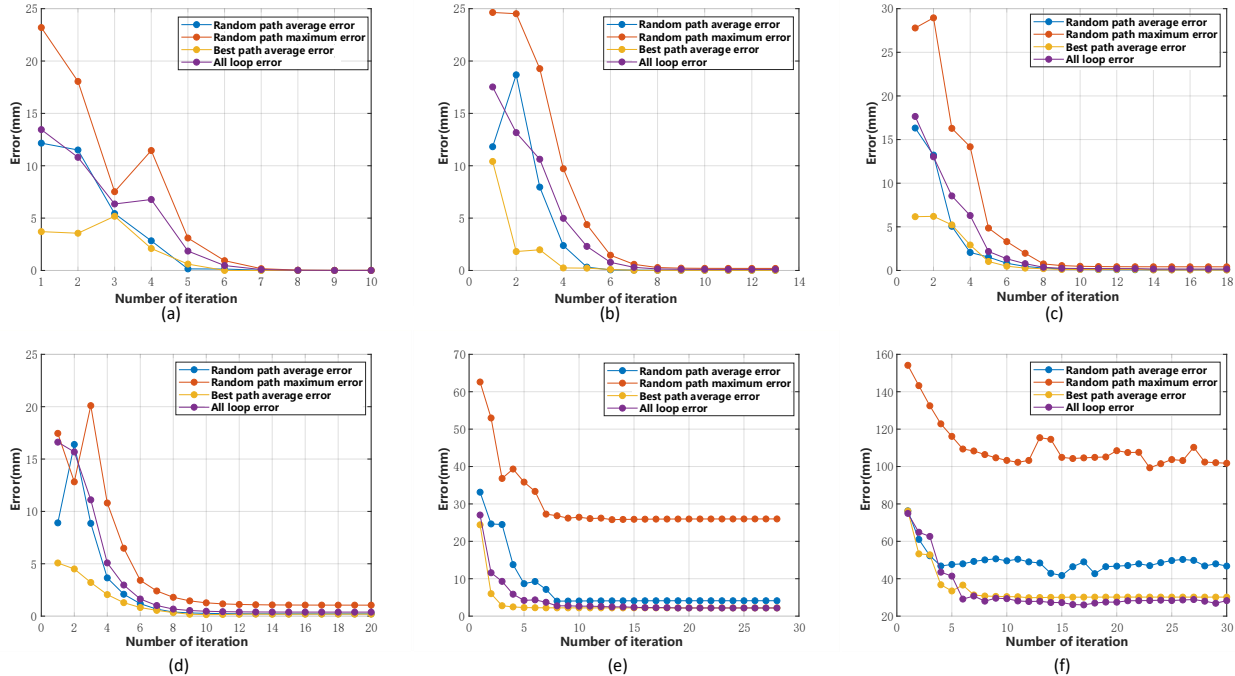

Fig. 4 The error variation comparison of models with different noise levels and the number of iterations. (a) $\sigma_{Noise} = 0.1mm$ (b) $\sigma_{Noise} = 0.2mm$ (c) $\sigma_{Noise} = 0.5mm$ (d) $\sigma_{Noise} = 1mm$ (e) $\sigma_{Noise} = 10mm$ (f) $\sigma_{Noise} = 100mm$

TABLE III
THE ERROR COMPARISON OF MODELS WITH DIFFERENT NOISE LEVELS

| Noise Level (mm) | VRD path Calibration Error (Our Method) | Average Random Path Calibration Error | Worst Random Path Calibration Error | All Loop Calibration Error |
|---|---|---|---|---|
| $\sigma_{Noise} = 0.1$ | 0.0019 | 0.0021 | 0.0136 | 0.0017 |
| $\sigma_{Noise} = 0.2$ | 0.0092 | 0.0217 | 0.1948 | 0.0092 |
| $\sigma_{Noise} = 0.5$ | 0.0516 | 0.0845 | 0.4185 | 0.0527 |
| $\sigma_{Noise} = 1$ | 0.1895 | 0.2167 | 1.0485 | 0.1836 |
| $\sigma_{Noise} = 10$ | 2.2018 | 4.1173 | 25.9876 | 2.1764 |
| $\sigma_{Noise} = 100$ | 30.1567 | 42.5941 | 99.5153 | 28.2285 |

## B. Robot Experiments

In order to validate the effectiveness of the general calibration method for MEMHS, a real robot experimental platform shown in Fig. 6 has been built. The experiment platform is composed of two UR5 robots (6-DOF, ±0.03mm repeatability), one UR3 robot (6-DOF, ±0.03mm repeatability), and five Hikvision cameras (MV-CE060-10UC). Three cameras are mounted at the end of the three robotic arms, and the other two cameras are placed in a fixed position within the workspace. The two cameras have no common field of view, and the intrinsic parameters of these cameras have been calibrated using the method proposed in [8]. Firstly, the calibration experiment is carried out using the MEMHS calibration proposed method, then the reprojection error is used to verify the calibration accuracy. Secondly, as a comparative experiment, Wu's dual robotic arms calibration method [18] is used to sequentially calibrate two robotic arms respectively. Finally, a graphical demonstration experiment can more intuitively illustrate the effectiveness of the proposed algorithm.

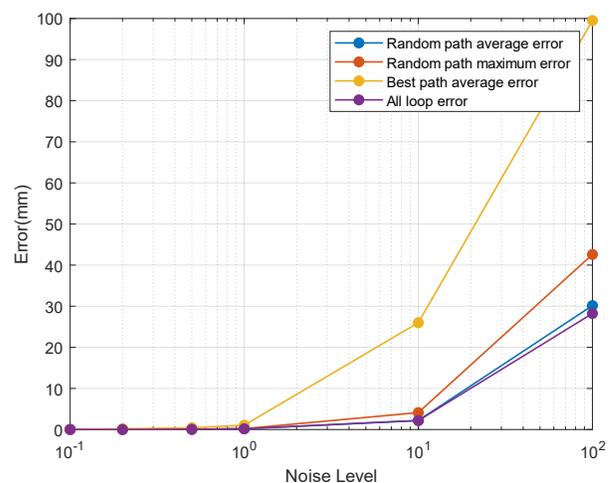

Fig. 5 Impact of the measurement noise.

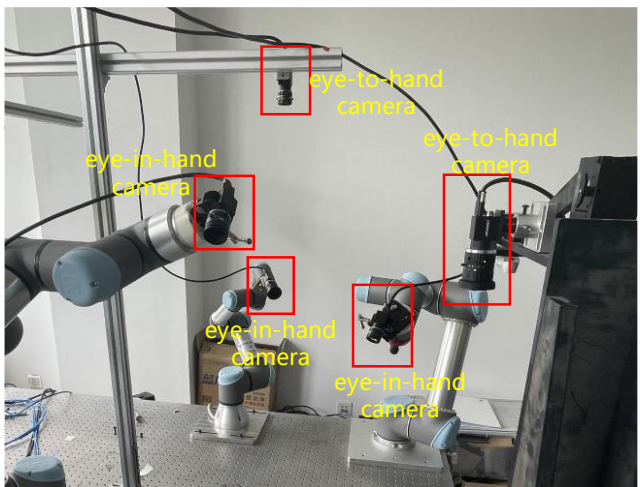

Fig. 6 Five-eye three-hand MEMHS.

### 1) Calibration Accuracy Verification

The MEMHS can be represented by the following model in Fig. 7. The solid lines represent measurable edges. In this MEMHS calibration experiment, we mainly focus on the transformation matrices corresponding to the red dash lines. These matrices cannot be measured directly but can be obtained by transferring values from nearby matrices, so these calibration results can effectively test the proposed calibration method. The weights of each vertex are shown in Fig. 7. For camera vertices, the weights are selected according to the camera's FoV, pixel density, and keypoint reprojection error during the intrinsic calibration process. For robot vertices, the weights are selected according to the reference manual from manufacturers.

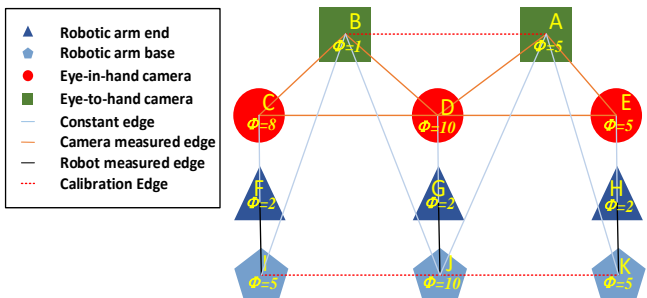

Fig. 7 The MCG of experiment MEMHS.

We first obtain the optimal calibration path according to the MCG. The calibration starts from D. We generate 30 configurations for each edge in the optimal calibration tree. Record the camera measurement result and robot joint parameters to perform initial calibration with hand-eye calibration [10]. A calibration board with $11 \times 8$ keypoints and $35mm$ spacing is placed in the center of the workspace to obtain edges between D-B, D-A, D-C, and D-E. Then smaller calibration boards with $11 \times 8$ keypoints and $25mm$ spacing are mounted at each robot end to obtain edges between C-J, D-I, and D-K. Finally, the edge J-G, I-F, and K-H is obtained with the robots' kinematic model. Since the true value is unknown in this experiment, the closed-loop error is calculated to measure the accuracy. The overall mean error in the optimal calibration loop is $1.1081mm$.

Then, the closed-loop optimization is conducted on the optimal calibration loop. With 48 iterations, the overall closed-loop mean error drops to $0.6182\ mm$, which is shown in Fig. 8.

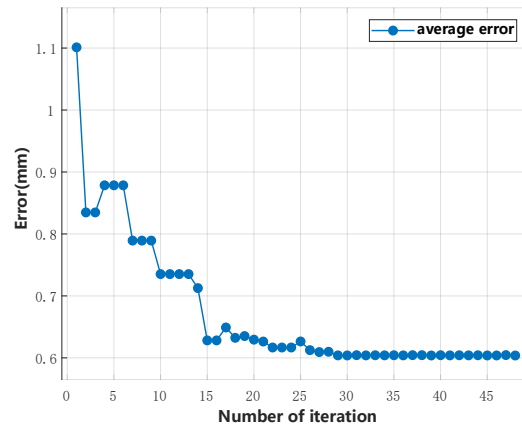

Fig. 8 Error variation during iteration in the experiment

### 2) Comparative Experiment with Wu's Method

To further illustrate the effectiveness of our proposed method in a multi-robot scenario, Wu's method [18] is used to calibrate these robots separately in the same experimental condition. The three robots are split into two dual robot sets, then perform calibration. The detailed calibration process is as follows.

1. Use Wu's method to calibrate the UR3 robot (vertex J-G-D) and one of the UR5 robots (vertex K-H-E).
2. Then also use Wu's method to calibrate the UR3 (vertex J-G-D) robot and the other UR5 robots (vertex I-F-C).
3. A calibration board is placed in the common field of view of the eye-to-hand camera (vertex A) and the UR3 robotic arm's eye-in-hand camera (vertex D) to determine the pose between vertex A and vertex D.
4. Repeat step 3 with vertices D and B.
5. Measure poses of vertices C-E with a calibration board placed in the common field of view.
6. The remaining edges to be calibrated are computed by pose translation between vertices.

The calibration results using this method are as follows. The difference from the calibration method proposed in this paper is about 0.4787mm and 0.8852mm for ${}^I T_J$ and ${}^J T_K$ respectively.

Since the true value of this experiment cannot be measured, we compare the calibration results by verifying the closed-loop error as shown in TABLE IV.

TABLE IV
CLOSED-LOOP ERROR BETWEEN OUR METHOD AND WU'S METHOD

| Closed loops | Our Method (mm) | Wu's Method (mm) |
| --- | --- | --- |
| $C - F - I - J - G - D$ | 0.6274 | 0.4213 |
| $D - E - H - K - J - G$ | 0.6345 | 0.5812 |
| $C - E - H - K - J - I - F$ | 0.6938 | 1.5923 |

Although Wu's dual robots' calibration method performs well in the situation of two robots, our method has more calibration accuracy in the multi-robot scenario, especially in non-direct optimized loops. The following trajectory drawing experiment can provide a more intuitive demonstration of the calibration accuracy.

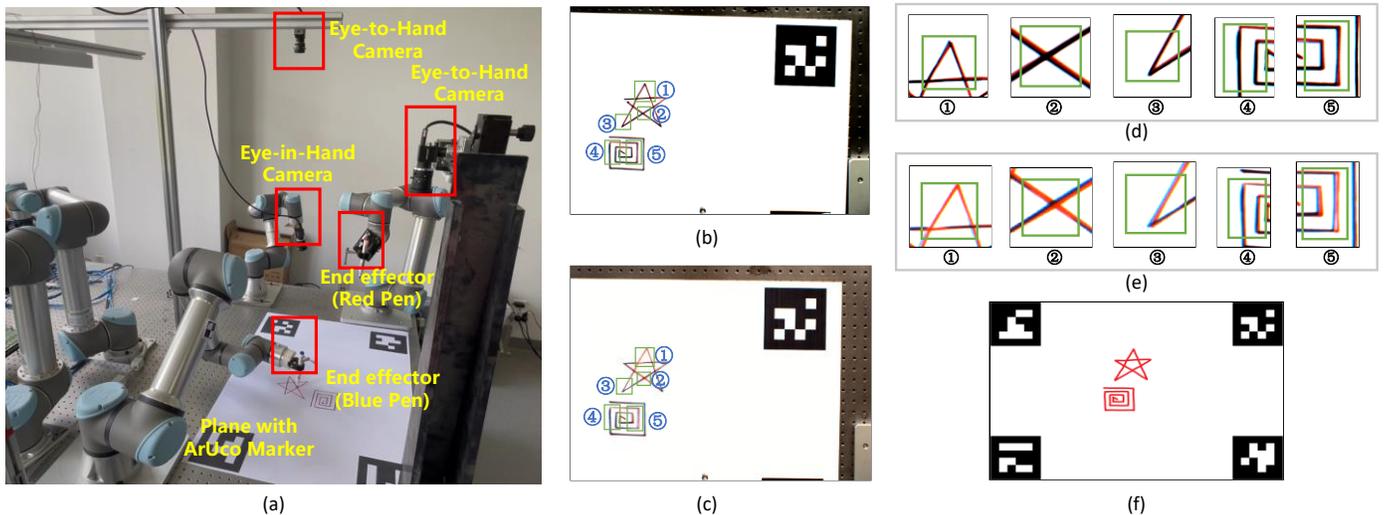

Fig. 9 (a) The MEMHS used in the drawing experiment. (b) Actual drawing trajectory with calibration result using our method. (c) Actual drawing trajectory with calibration result using Wu's method. (d) Trajectory and its details drawn by our method. (e) Trajectory and its details drawn by Wu's method. (f) The desired trajectory on the computer.

### 3) Vision Guided Drawing Experiment

In this drawing experiment, the calibration result of both our method and Wu's method mentioned above is used to test the calibration accuracy. Remove the cameras at the ends of the two robots and install markers to simulate tools as shown in Fig. 9(a). Use the two fixed cameras and one robot to guide the other two robotic arms to write on the plane. A plane printed with ArUco markers is placed within the range of the fixed camera and one robot. These cameras measure the marker pattern through the PnP [27] algorithm to detect the pose and size of the plane. Draw trajectories on the computer shown in Fig. 9(f), and then the robotic arms will draw the trajectories on the plane in turn with our calibration result and Wu's calibration result. The trajectories drawn by the robotic arms overlap each other and are the same as the preset trajectories shown in Fig. 9(b-e). Both trajectories drawn by our method and Wu's method have the same pattern, but the trajectories drawn by our method have a higher degree of consistency. With our method, more parts of the first drawn red trajectory are covered by the latter drawn black trajectory and has fewer trajectories' inconsistent areas compared with Wu's method.

## V. CONCLUSION

In this paper, a general calibration method for MEMHS is proposed, which is the fundamental of collaboration of multiple camera robot systems. Our proposed calibration method includes two steps. In the first step, the minimum spanning tree based on system settings is used to obtain the optimal calibration order. In the second step, a numerical optimization solution based on the closed-loop error of the minimum calibration loop is proposed to optimize the poses between the units. The proposed calibration method achieves calibration tasks without additional high-precision sensors. Simulation and robot experiments strongly demonstrate the effectiveness of the proposed method. However, the influence of different system settings and the kinematics parameters of the robotic arm itself requires further research. Furthermore, how different sensor types, such as omnidirectional cameras, affect the calibration accuracy should also be investigated. All these issues require further study. This method is a general calibration method that can be applied to different settings of camera robot systems. Such as high precision multi-robot Aero-engine superalloy repair and grinding, multi-robot integrated circuit micro-assembly system, multi-robot collaborative welding of automotive body, and other scenes requiring precise cooperation of multiple robots. The calibration algorithm of the multi-hand-eye robot proposed in this paper can be widely used in the above scenarios.

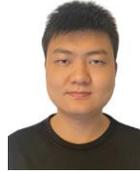
**Zishun Zhou** received the B.E. degree from Harbin Institute of Technology, Harbin, China, in 2020. He is currently a postgraduate student in electrical engineering of the Australian National University, Canberra, Australia. And currently work as a student intern under the supervision of Xilong Liu in the Institute of Automation, Chinese Academy of Sciences, Beijing, China. His research interests include visual measurement, multi-view stereo and 3D dense reconstruction.

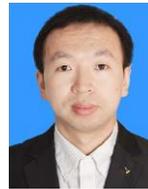
**Liping Ma** received the B.S. degree from Hunan Agricultural University, Changsha, China, in 2007, and the Ph.D. degree in mechanical engineering from the Beijing Institute of Technology, Beijing, China, in 2015. He is currently an Associate Professor with the Research Center of Precision Sensing and Control, IACAS. He is also with the School of Artificial Intelligence, UCAS. His current research interests include robot design, electromechanical system design, visual measurement, and application.

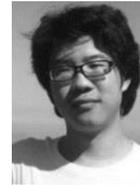
**Xilong Liu** received the B.S. degree from Beijing Jiaotong University, Beijing, China, in 2009 and the Ph.D. degree in control theory and control engineering from the Institute of Automation, Chinese Academy of Sciences, Beijing, in 2014. He is currently an Associate Professor with the Research Center of Precision Sensing and Control, Institute of Automation, Chinese Academy of Sciences. His current research interests include image processing, pattern recognition, and visual measurement.

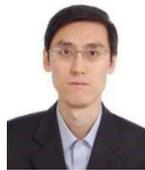
**Zhiqiang Cao** received the B.S. and M.S. degrees from Shandong University of Technology, Jinan, China, in 1996 and 1999, respectively. In 2002, he received the Ph.D. degree in control theory and control engineering from the Institute of Automation, Chinese Academy of Sciences, Beijing, China. He is currently a Professor in the State Key Laboratory of Management and Control for Complex Systems, Institute of Automation, Chinese Academy of Sciences. His research interests include service robot and multirobot coordination.

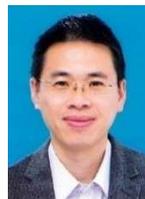
**Junzhi Yu** received the B.E. degree in safety engineering and the M.E. degree in precision instruments and mechanology from the North University of China, Taiyuan, China, in 1998 and 2001, respectively, and the Ph.D. degree in control theory and control engineering from the Institute of Automation, Chinese Academy of Sciences, Beijing, China, in 2003. From 2004 to 2006, he was a Post-Doctoral Research Fellow with the Center for Systems and Control, Peking University. He was an Associate Professor with the Institute of Automation, Chinese Academy of Sciences, in 2006, where he was a Full Professor in 2012. In 2018, he joined the College of Engineering, Peking University, as a Tenured Full Professor. His current research interests include intelligent robots, motion control, and intelligent mechatronic systems.